\UseRawInputEncoding

\documentclass[11pt]{article}
\usepackage{amsmath}
\usepackage{a4wide}
\usepackage[table]{xcolor}
\usepackage{todonotes}
\usepackage{changes}
\usepackage{amsfonts}
\usepackage{cuted}
\usepackage[breaklinks]{hyperref}

\usepackage{authblk}

\definechangesauthor[name=Luke, color=red]{LD}
\definechangesauthor[name=Rob, color=green]{RM}

\newcommand{\lukehidden}[1]{}

\usepackage{setspace}
\usepackage{titlesec}
\titlespacing*{\section}{0pt}{0.5\baselineskip}{0.2\baselineskip}
\titlespacing*{\subsection}{0pt}{0.5\baselineskip}{0.2\baselineskip}

\title{Repurposing of Resources: from Everyday Problem Solving through to Crisis Management}

\author[1]{Antonis Bikakis}
\author[1]{Luke Dickens}
\author[2]{Anthony Hunter}
\author[1]{Rob Miller}

\affil[1]{Dept of Information Studies, University College London, London, UK}
\affil[2]{Dept of Computer Science, University College London, London, UK}

\begin{document}

\maketitle

\begin{abstract}
The human ability to repurpose objects and processes is universal, but it is not a well-understood aspect of human intelligence. Repurposing arises in everyday situations such as finding substitutes for missing ingredients when cooking, or for unavailable tools when doing DIY. It also arises in critical, unprecedented situations needing crisis management. After natural disasters and during wartime, people must repurpose the materials and processes available to make shelter, distribute food, etc. Repurposing is equally important in professional life (e.g. clinicians often repurpose medicines off-license) and in addressing societal challenges (e.g. finding new roles for waste products,). Despite the importance of repurposing, the topic has received little academic attention. 
By considering examples from a variety of domains such as every-day activities, drug repurposing and natural disasters, we identify some principle characteristics of the process and describe some technical challenges that would be involved in modelling and simulating it. We consider cases of both substitution, i.e. finding an alternative for a missing resource, and exploitation, i.e. identifying a new role for an existing resource. 
We argue that these ideas could be developed into general formal theory of repurposing, and that this could then lead to the development of AI methods based on commonsense reasoning, argumentation, ontological reasoning, and various machine learning methods, to develop tools to support repurposing in practice. 
\end{abstract}

\section{Introduction}

The human ability to repurpose objects and processes is universal, but not a well-understood aspect of human intelligence. Repurposing arises in everyday situations such as finding substitutes for missing cooking ingredients, through to critical, unprecedented situations needing crisis management. After natural disasters and during wartime, people must repurpose the materials and processes available to make shelter, distribute food, etc. Amid the COVID-19 crisis health provision has been adapted, with organisations and businesses transforming and repurposing themselves accordingly. At the domestic level, parents became teachers, kitchens became classrooms and cars became quiet spaces for making business calls. Repurposing is important in professional life (e.g. clinicians often repurpose medicines off-license) and in addressing societal challenges (e.g. a Thames Water plant in Slough near London for turning sewage into fertiliser
or NGOs helping Zanzibar women produce agar from seaweed).

There is a pressing need to better understand repurposing as an important way that humans cope with unexpected situations. The COVID-19 crisis has shown how as a society, we can adapt to each new problem that arises, and often this has involved repurposing, whether it is exhibition centres as hospitals, Kent motorways as parking area for trucks when they cannot pass through to France, the corticosteroid medication Dexamethasone as a treatment for patients with acute respiratory distress syndrome, etc. However, the COVID-19 crisis has also shown us that we could have benefited from adapting more quickly and more creatively. Indeed, there are some major problems that are not being adequately addressed, for example, an upsurge in loneliness, rising mental health issues, greater inequality, with this being compounded by children from poorer backgrounds losing out more from closed schools, etc. Some of these problems could potentially be tackled by repurposing. 

But we should not focus exclusively on COVID-19. There will be more unexpected problems coming to the fore whether from microbiology (antiseptic resistant bacteria, viruses that are more serious than COVID-19, etc.), from geology (earthquakes, volcanic eruptions, etc.), from weather (flooding, desertification, forest fire, etc.), and from war, as well as predictable but significant problems such as shortages of water, petroleum, land, food, precious metals for electronics, and the effects of global warming and pollution. In all these cases, repurposing is required. 

So there is no shortage for applications of the theory and technology for repurposing. In some cases, we anticipate that completely automated solutions can be obtained in the short-term (e.g. substitution of cooking ingredients, or tools in DIY). In other cases (e.g. recycling, or crisis management), we anticipate that only interactive solutions (where user input is required) could be developed in the short-term. 

Repurposing offers some exciting challenges for AI, and could offer some valuable case studies for natural language processing, machine learning, and knowledge representation and reasoning, in particular for commonsense reasoning. To investigate these challenges, we proceed by considering some key concepts in repurposing, reviewing the current literature on the topic, and then exploring some potential research questions for AI in repurposing.



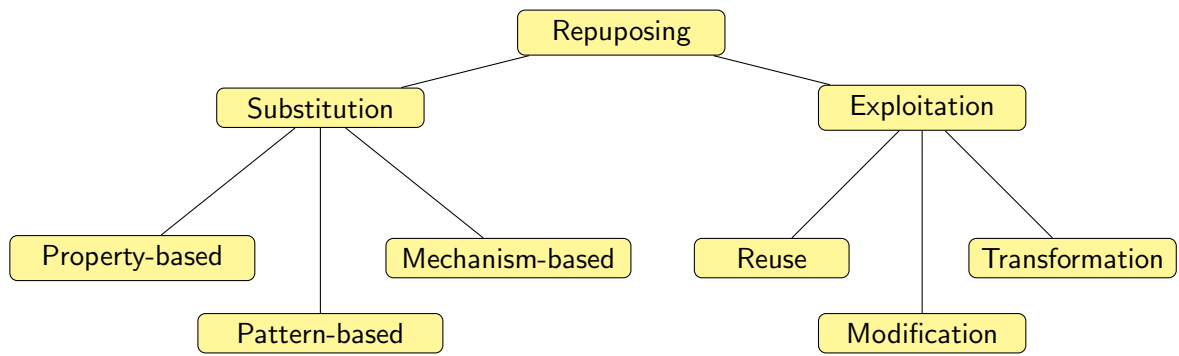
\begin{figure*}
\begin{center}
\begin{tikzpicture}[module/.style={rectangle,text centered,text width=18mm,fill=yellow!50,rounded corners=1mm,draw},
module2/.style={rectangle,text centered,text width=25mm,fill=yellow!50,rounded corners=1mm,draw},
module3/.style={rectangle,text centered,text width=30mm,fill=yellow!50,rounded corners=1mm,draw}]
\node[module2] (r) at (5,3) {Repuposing};  
\node[module2] (s) at (1,2) {Substitution};  
\node[module2] (e) at (9,2) {Exploitation};  
\node[module3] (s1) at (-1.5,0) {Property-based};  
\node[module3] (s2) at (1,-1) {Pattern-based};  
\node[module3] (s3) at (3.5,0) {Mechanism-based};  
\node[module] (e1) at (7,0) {Reuse};  
\node[module2] (e2) at (9,-1) {Modification};  
\node[module2] (e3) at (11,0) {Transformation};  
\path	(s) edge node[] {} (r);
\path	(e) edge node[] {} (r);
\path	(s1) edge node[] {} (s);
\path	(s2) edge node[] {} (s);
\path	(s3) edge node[] {} (s);
\path	(e1) edge node[] {} (e);
\path	(e2) edge node[] {} (e);
\path	(e3) edge node[] {} (e);
\end{tikzpicture}
\end{center}
\caption{Key concepts for repurposing.}
\label{fig:concepts}
\end{figure*}

\section{Concepts in Repurposing}

In this section we draw out some key concepts in repurposing. Figure~\ref{fig:concepts} presents an overview and a classification of these concepts.

\subsection{Substitution}

Substitution is a scenario commonly involving repurposing. We have a task (doing or making something) for which we lack a resource. We solve the problem by repurposing an alternative to substitute for the missing thing. 
For example, we could have a recipe for bread that includes the ingredient butter. In fact, olive oil can be used as a substitute. Someone unaware of this may nonetheless be able to work it out by reasoning about the relevant common properties of butter and olive oil. Both are used in baking to add fat, both make a relatively small change to the flavour, and both are neutral in their savoury vs sweet impact. In order to mimic this capability with scalable computational reasoning, we have various options. We briefly discuss three analytical approaches here and discuss supporting technologies.
\lukehidden{Does substitution potentially include doing without said ingredient? Substituting with nothing.} 

\begin{description}

\item[Pattern analysis] involves looking for patterns in observed practices that we can use as indicators of substitutability. 
For example, olive oil and butter can be regarded as being in a paradigmatic relation (in a cooking context) since they often appear in the same role in many recipes. This idea can be used in tools to help families discover similar recipes and, with appropriate models, may uncover unobserved substitutions by virtue of transitive relationships, e.g. olive oil substituting butter, sunflower-oil substituting olive oil. However, it cannot alone uncover more creative or `make do' substitutions not in existing recipes. Also, this data-driven approach does not provide explanations for why particular substitutions work. \lukehidden{ These sorts of techniques could be carried out on free-text datasets or on more structured data too, e.g. databases of recipes where each ingredient had a UID. Equally, linguistic analysis could also describe techniques for mining for properties or justifications of substitution, e.g. with regular expressions to identify sentences like: "If you don't have any X you can replace this with Y" or "Some people prefer to use Z".}

\item[Property analysis] involves discovering, collecting and systematising knowledge about the relevant properties of substituted items. For example, looking at some general functional properties of foods (e.g. from \url{www.ifst.org}) such as aerates, binds, thickens and glazes, we see that butter and olive oil have the role of \emph{binding} in baking, and both bring about similar structural properties (plasticity and softness) and similar flavour properties. Using a cooking ontology (e.g. WhatToMake (\url{purl.org/heals/foodon/})), we could model recipes, ingredients, ingredient properties, cooking processes and relationships among these, and reason about missing properties, more abstract concepts and higher order relationships such as a hierarchy of classes of ingredient properties, which could lead to more creative/surprising substitutions. The advantage of property analysis over pattern analysis is that relationships are justified in terms of explanatory factors, and so proposed substitutions can be explained and interrogated, and errors or omissions exposed and corrected.  However, the properties that are relevant for a substitution, are often dictated by the context of the task. For example, in place of flour, agar can thicken a sauce, but cannot be used to prevent dough sticking. Capturing and reasoning with this context-sensitive knowledge requires a semantically rich representation of the repurposing task, which does not only include the items required to perform the task, but also the properties of these items that make them suitable and their dependencies in the specific task.  

\item[Mechanistic analysis] involves modelling the underlying mechanisms of the process for which a part is being substituted. For this modelling, we require a clear and explicit representation of the knowledge involved. For instance, a recipe can be represented by a logical model where each step, action or process, such as adding an ingredient, boiling, baking, etc, brings about a change. For making bread, a model would involve starting with an empty bowl, adding yeast, flour, etc, and mixing to obtain a dough, followed by other processes such as kneading and baking.  So the pre- and post-conditions of each step can be modelled to give a sequence of states and identify action-effect relationships. 
An advantage over property analysis is that richer knowledge about the roles of substitutions is represented, exposing relevant properties more clearly, and informing consequences and mitigations for given substitutions.
e.g. sour-dough starter may substitute for yeast changing the flavour and requiring longer rising time, and brown pasta substituting white pasta increases fibre content but requires a longer cooking time. 
How each component works with other components in the context of the actions applied is captured, and so potentially better and more creative substitutions can be identified by reasoning with the logical models.

\end{description}

The three analyses: pattern, property and mechanistic, are presented in order of richness of representation. All can be used to find substitutes for missing ingredients or components. 
Our examples have focused on food, but substitution occurs in many situations, e.g.\ missing tools in DIY, or missing resources (shelter, communication networks, food, etc.) after a natural disaster.

\subsection{Exploitation}

We consider exploitation to be a key concept in repurposing. It involves two aspects: {\bf Exploitables} which are the objects, material, substances or other resources that we have in excess and that we would like to better use (exploit); and {\bf demands} which are open problems that require (and possibly lack) some objects, material, substances or other resources. As such, reasoning about exploitation has clear benefits for sustainability and waste reduction, e.g. automatically generating recipes to make use of surplus food.
The analyses discussed for substitution are also applicable to exploitation. However, matching exploitables with demands may have the additional challenge that they are described in different functional terms. Our exploitable might be a plastic bottle described as a \emph{soft drink container}, whereas our demand might be for \emph{carrying water}. Repurposing is therefore needed to match an exploitable with a demand. 
Given an exploitable, we next identify three modes of exploitation via repurposing.

\begin{itemize}

\item In {\bf reuse}, we can keep the exploitable intact, and use it as is for a different but related purpose. A plastic bottle can be repurposed as a water bottle for a hiking trip, because the properties of soft drinks and water are similar. Both are liquid comestibles needing to be carried in a water-tight container. 

\item In {\bf modification}, we make changes to the exploitable to use it for a new purpose. The processes of change will typically be mechanical and will result in limited changes that may be described in terms of the property analysis outlined earlier. For example, modifications to our plastic bottle may range from the relatively minor (e.g.\ cutting off the top to make a cup, or cutting off the top and putting some holes in the bottom to make a plant pot) 
to the more substantial
(e.g.\ cutting top, bottom and length to create a sheet of plastic for drain-pipe repair). But in all cases key properties of the water bottle are maintained including its ability to completely or partially contain liquid.


\item In {\bf transformation}, we make more substantial changes to the exploitable that make it into a ``new" object/substance. A plastic bottle can be shredded, melted down and formed into small pellets as raw material for many different products (e.g.\ garden furniture, fibre-pile clothes). The resulting object can share relatively few properties with the original plastic bottle, which was hollow, rigid, transparent and liquid-containing. For instance, a fibre-pile pullover is soft, opaque, foldable and cannot contain liquid. Nonetheless, other properties are retained, e.g.\ thermally insulating, allowing us to reason about alternative uses.
\lukehidden{Removed: In order to model transformations, we need to capture sequences of more complex mechanical, thermal, and chemical processes.}
\lukehidden{The distinction between radical modification and transformation is a little vague here. }

\end{itemize}

For physical objects, modifications capture how the items can be changed from one state to another. For example, how tools can be used to cut pieces off, add holes, smooth or roughen surfaces, etc, and how pieces can be added such as brackets, hinges, supports, etc, with glue, nails, screws, and bolts, or by using welding, soldering, etc. Ideally, these modifications can be seen as reflecting what would be available from a well-equipped engineering workshop. So each kind of modification takes an object in a particular state, and a tool, in order to change the state of that object. This kind of state-based reasoning applies to diverse objects including electronic devices, furniture, motor vehicles, machinery, and buildings. 
For non-physical objects, or for processes, modifications are potentially more subtle and wide-ranging. For instance, for software code, modifications could be various kinds of tweak or patches or extensions.

Transformations call for more fundamental changes of state. For a physical object or a material, a transformation could be a physical, thermal, or chemical process that substantially changes the object or material. Consider for example manufacturing processes that take raw materials such as trees, and subject them to physical processes to chop and mash them, and then subject the resulting material to chemical processes that turn the wood eventually into paper. We can decompose these processes, and regard each like a tool that takes the input to change its state to the output. The sequences of these processes can be composed to capture how materials can be transformed into radically different products. 

In order to build a system for exploitation, we require knowledge of both the potential exploitables and possible demands. Both can be partially described in terms of properties (functional, mechanical, chemical, aesthetic, etc.), and these can be employed in a match-making exercise to assign one to the other. However, we also require knowledge of possible modifications and transformation steps. Then the process of repurposing is exploring this knowledge, either working from exploitables to demands in a data-driven way, or working from demands to exploitables in a goal-directed way, or combining these two reasoning modes. In exploring this space, a repurposing system needs to identify a sequence of steps that will take the exploitable to the demand. 
To facilitate this, the space of possible modification and transformation steps (mechanical, chemical, etc.) can be effectively formalised in terms of pre- and post-conditions, or cause and effect relationships. 
Comprehensive modelling would need to take into account the availability of  exploitables (quantify, location, cost), cost and availability of the transformation steps, and the value of and need (``market") for the demand.


\section{Domains for Repurposing}

In this section, we consider some domains and examples to further indicate the ubiquity of repurposing. We highlight some common themes and also discuss some interesting differences. We start with the cooking domain as it is both rich in types of repurposing and it is a helpful domain for understanding these ideas. We then consider some further everyday domains including gardening, DIY, electronic devices, and recycling. We follow this with drug repurposing, which is an increasingly important topic in healthcare, crisis management, and helping disadvantaged people.

\subsection{Cooking}

The domain of cooking is a very rich one for repurposing. Here we define cooking in a general food preparation sense -- even if no food is actually \emph{cooked} by heating. There are a number of reasons that may explain why cooking presents so many opportunities for, and examples of, repurposing. To begin, cooking is something that we all do, and we must all do, every day in order to stay alive, as such more cooking happens everyday around the world than almost any other non-autonomic activity. More than this, cooking is a unique blend of what is necessary and what is enjoyable, and as such has both practical and hedonic (what is un/pleasant) constraints. Also, most food-stuffs are perishable, meaning that they are at their best, both nutritionally and in terms of enjoyment, for a limited period of time. Moreover, food production is inherently variable both in yield and quality. This can be due to predictable factors, such as the seasonal cycle, and less predictable factors, such as rainfall and temperature variation. Fluctuations in the supply chain and demand adds additional uncertainty. 

These factors alone would provide fertile ground for both aspects of repurposing. In particular, variation in the availability of ingredients means that to make a dish we like, we may sometimes have to \emph{substitute} a preferred ingredient that is unavailable, for a suitable one that is available (see \textbf{substitution due to scarcity} below). On the other hand, there are times of plenty, when even our most favoured food-stuffs are present in such quantities that we must be creative in finding uses for them, i.e. a glut requiring \emph{exploitation} (see \textbf{exploitation of a glut} below). 

While these are probably the most straightforward reasons for repurposing in cooking, there are others. These can again be related to the unique relationship humans have with food. We often crave the familiar, as captured by phrases like \emph{chicken soup for the soul} or \emph{Momma's apple pie}. Nevertheless, we value novelty too, and this hedonic impulse leads us to continually try new cuisines, new restaurants, new dishes and new versions of old favourites. Moreover, we often view food as sitting within some culinary context. For instance, many don't like mixing sweet with savoury, and it is common to pair dishes from the same culinary tradition\footnote{It should be noted that some, such as fusion restaurants, make a virtue of mixing food from different cultures and this again speaks to the complexity of proposing substitutions.}, e.g. a greek salad with moussaka. We argue that it is this tension between the familiar and the novel that makes the cooking domain so interesting for repurposing, and outline a small number of, perhaps less obvious, opportunities for repurposing. These include: substitution due to dietary constraints;  exploitation around a theme; and repurposing a process. We expand on each of these situations in their own paragraphs below.

\subsubsection{Substitution of ingredients due to scarcity}

Scarcity, arising from fluctuations in supply, price and quality can mean that a planned or desired dish cannot be made exactly as intended. This either means that you do without the dish, that you make it with an overpriced or poor quality ingredient, or that you substitute that ingredient for another. This last option is such a common scenario that many recipes include notes on reasonable substitutions and some recipe websites have even taken to asking users to propose their own ideas about substitutions as a form of crowdsourcing \footnote{For instance, Sainsbury's supermarket chain provide recipe suggestions on \href{https://recipes.sainsburys.co.uk/recipes}{their website}. On each recipe, they have the following text with a link: "Don't have the ingredients or just fancy a change? Here's some ideas", as well as a "Tell us your swap" button to crowdsource ingredient substitutions.}. The ability to successfully predict ingredient substitutions has commercial application too. For instance, a number of home delivery supermarkets now automatically substitute unavailable items from grocery orders (although not always successfully). Ingredient substitution is a challenging problem and isn't just about the properties of ingredients, as Brian O'Driscoll said: ``Knowledge is knowing that a tomato is a fruit. Wisdom is knowing not to put it in a fruit salad.''

Proposing substitutions for a given ingredient in a known recipe is a form of recommendation, but with additional complexity to other, more conventional recommender systems, e.g.  for films, and this is itself an imperfect science. Film recommender systems typically offer a list of recommendations from which the user can select, and so only require a high chance that the user will find a reasonable suggestion within a list of $K$ items that the user can browse through\footnote{Evaluations for recommender systems in the literature reflect this use case and typically measure recommender performance with such metrics as hits@K (i.e. how many good answers appear on average in a list of $K$ recommendations.}, unlike the online grocery store that has just one opportunity to propose an appropriate substitution.

Even in food recommendation systems with a browsing feature, there are other additional complexities. Importantly, the recommendation for food substitution must not only satisfy the recipient's preferences but also work well in the context of the dish. Having said this, the other ingredients in the dish may provide important information about the user's current preferences\footnote{In the case of the grocery order substitution, there is the added complexity that the intended dish is unknown and must be inferred from the rest of the grocery order.}. Finally, the use of a different ingredient, may require different processing requirements. To fully encompass/reason about the implications of a proposed substitution, an account of how this changes the recipe method is also needed. For instance, replacing white flour with brown in bread making can lead to longer rising and cooking times.

\subsubsection{Evidence of ingredient substitution}

Although, ingredient substitution due to scarcity is a challenging prediction task, there are reasons to believe it is achievable. Substitutions happen all the time in the domain of cooking and so there is a rich vein of information to be mined here. Recipe books contain substitution suggestions, as mentioned, but (some books) also contain related information about how to reason more generally about substitutions. There is also a wealth of recipe suggestions shared online, both by interested individuals and larger organisations. Taken as a whole, this (these) extended corpus (corpora) of recipes contain many similar recipes, with similar names and/or similar ingredients. The commonalities and differences between recipes give important information about what ingredients work together (and in what contexts) as well as what doesn't (or maybe just hasn't been tried). 

These similarities can be seen at different levels of abstraction. At the time of writing, the phrase "spaghetti Bolognaise recipe" gave over $600'000$ search results on a web-search. Inevitably, these recipes will contain differences in: bulk ingredients (e.g. beef versus pork mince), flavourings (e.g. basil versus mixed herbs), method (e.g. different cooking times) and so on. Nonetheless most recipes will appear somewhat familiar (even if they might scandalise an Italian). It may be possible to capture and exploit this information by just looking at ingredient lists for a large corpus of recipes and appropriately clustering/partitioning the space of ingredients and/or recipes.

Digging down to look at components of these Bolognaise recipes, we may notice that many recipes involve first slowly saut\'{e}ing finely diced onions, celery and carrots. This component is referred to as \emph{soffritto} by the Italians, and appears in other dishes too. Any suitable substitution in a soffritto may therefore apply more generally than to just one dish. What is more, there is good reason to believe that substitutions in soffritto are possible. This or similar rich vegetable savoury bases are made in other cultures too, with different names, and sometimes, slightly different ingredients. For instance: the French make mirepoix in the same way; the Spanish \emph{sofrito} but may use garlic, onions, peppers and tomatoes; while the Germans may use carrots, celery root, leeks and parsley to make \emph{Suppengr\:{u}n}. Outside of Europe, Dominican, Puerto Rican and Creole cooking have very similar bases called saz\'{o}n, recaito and the Holy Trinity respectively. How to capture evidence for parts of a dish presents a greater data-processing challenge. Such descriptions may not appear readily in individual ingredient lists (unless explicitly structured as such). There may be information in the food preparation method, but this requires advanced natural language processing (NLP)  techniques to translate from natural language to machine-interpretable knowledge.

Some of the evidence described above is more explicitly informed by the need for substitution but other parts are more implicitly motivated or merely capture general principles of what goes well with what. Together it represents a huge and valuable resource for reasoning about this domain though, and some have already begun to exploit and formalise this knowledge (for instance the knowledge base provided by \url{foodkg.github.io/} and Food Substitutes \url{www.foodsubs.com}).

\subsubsection{Exploitation of a glut}

For those that grow their own vegetables, most will have experienced a glut, where a particular food-stuff is available in very large quantities, but, unprocessed, has a short lifespan. This may then require creative ways to make good use out of the resource, and identifying this represents a demand for exploitation. This exploitation may well involve extending the useful life of the food-stuff perhaps through pickling, curing, freezing, drying or other preservation technique. Such techniques can substantially affect other properties (taste, texture, nutritional content) of the ingredient, and so this use case may involve being able to reason about exploitation requiring modification or transformation. Another way to use up large quantities of a food-stuff is to think about creative ways to use a given ingredient. We discuss this in more detail under \textbf{exploitation around a theme}.

\subsubsection{Substitution due to dietary constraints} 

Other than scarcity, another reason to replace one or more ingredients with appropriate substitutes in a dish is due to dietary constraints. The substituting ingredient here must satisfy competing demands: it must retain some properties of the substituted ingredient, omit others, and satisfy the preferences of the consumer. In order to make predictions here a detailed understanding of the properties of ingredients is needed, such as is captured in food knowledge graphs mentioned above. Some constraints will be hard/strict, particularly where they relate to such things as allergies. Other constraints will be softer, for instance, a vegetarian would not want their chips to be deep fried in lard (smoke point $188^\circ C$), and a suitable substitute may demand a similar \emph{smoke point}, but a number of substitutes are recommended under these circumstances each with slightly different smoke points, e.g. sunflower oil (smoke point $230^\circ C$). 
As this may be a safety critical objective this also has implications about how substitution predictions are integrated with human processes. For instance, a food safety qualified individual may have to mediate between an automated decision-support system and the decision to substitute one ingredient for another. As a consequence, the reasoning/prediction process may need to be explained/interrogated and the underlying knowledge base or model edited/curated.

\subsubsection{Exploitation around a theme}

This use case encompasses situations in which one or more ingredients are selected as the basis for a dish, but with a certain freedom to creatively use that ingredient combination. As discussed, this can emerge as part of exploitation of a glut, but also applies more widely, for instance if a restaurant chef would like to add diversity to a menu, or when a home cook wants to prepare something special for a friend/family member based around known preferences. There is a particular emphasis on creativity here, and so a prediction system may benefit from supporting less obvious/well-evidenced proposals. As such, supporting technologies may need to support control over the creativity of a proposal. In machine learning prediction terms this may be described as a greater risk appetite and so relate to a different prediction threshold.

\subsubsection{Repurposing a process} 

Up to this point, we have only described food-based repurposing opportunities in terms of replacing or finding uses for ingredients. Here we briefly discuss the idea that a process or method described for one (a set of) ingredient(s) can be applied (as is or in an adapted form) to another. The result may be a modified ingredient which will then need some creativity to place within a dish (exploitation). For example, freezing is a common preservation technique to extend the life of some ingredients, e.g. cuts of meat, carrots, peas, but is often overlooked for others, e.g. avocados, mangoes, tomato. Frozen avocados can be defrosted and used in a dish such as guacamole (which may not be much different from how the fresh ingredient would be used). However, frozen mangoes may be more appealing when transformed into mango sorbet.

\subsection{Gardening}

Consider a gardener with a plant with a long narrow stem (example adapted from \cite{Davis1998}). First, the gardener reasons that if it is windy, the stem might snap and therefore kill it. Then, she reasons that a stake could be used to give support. She doesn't have a stake, but she sees a nice straight branch on the ground that is a little bit taller than the plant. She pushes this into the ground near the plant. Next, she needs to use this as an impromptu support for the plant. She happens to have a hammer and nail in her pocket, but she reasons that if she nails the plant to the stake, it will damage the plant. She looks in her other pocket and finds an old crisp packet which she reasons she can use to loosely tie the plant to the stake. The gardener has therefore used various reasoning processes including commonsense scenario analysis to identify a risk, commonsense problem solving to tackle the risk, and this in turn called on commonsense problem solving to tackle the lack of resources. This has drawn on commonsense knowledge about plants, the effect of wind, objects that can act as a stake and string, and how to use them to mitigate the risk.

\subsection{DIY}
Plenty of ideas for repurposing household items are presented on websites such as \url{diyncrafts.com}. For example, repurposing broken chairs as hangers, benches or lawn swings, turning a broken washing machine drum into a coffee table or a stool, turning a broken suitcase into a table, pet bed or medicine cabinet, making a kitchen utensil holder from a broken rake, making an aquarium from an old TV or a broken computer monitor, etc. A common characteristic of all these examples is that they involve objects that can no longer be used for their original purpose but can be repurposed to meet other needs. They, therefore, fall in the category of exploitation. All these repurposing examples require some form of modification of the exploitables. For example, to repurpose a broken chair as hanger, one needs to to remove the back of the chair and then add the hanger hardware, while to turn an old TV into an aquarium one needs to remove the electronic pieces and then install the glass and the other aquarium parts. They therefore involve reasoning not only about the properties of the original object, but also about the changes of its state during the modification process.

\subsection{Electronic devices}

An interesting case study of repurposing of used smartphones was presented in \cite{zink14}, in which an app was created for two different smartphones to function as purpose-built personal in-car parking meters. The aim of the study was to investigate the environmental impact of smartphone repurposing as compared to traditional refurbishing, and the results showed that the option with the lower environmental impact would be to repurpose the smartphones. This case study is an example of exploitation, as the smartphone can no longer be used as a phone and alternative purposes are sought. Enabling such new types of use of smartphones or other computing devices does not require any modification to its hardware. The challenge, here, is to reason about the computational capabilities of the device and to write the software that enables the new functionality.

\subsection{Recycling}

There is much interest in recycling as there is wide-spread concern about ecological issues (e.g. sustainability of human activity, global warming, biodiversity, and depletion of global resources). 
Yet recycling of products and materials has long been an important kind of economic activity. Often, this can be about turning waste into raw materials that can be used again for the same or similar purpose. 
A good example is the collection of aluminium soft drinks cans that are melted down to make aluminium sheets that can then be used for new aluminium products including soft drinks cans again. Producing aluminium from bauxite is an expensive and environmentally damaging activity. It uses enormous amounts of energy, whereas smelting recycled aluminium uses relatively little energy. 

There are plenty of other examples of turning waste into raw materials including steel from food cans, car bodies, ships, building materials, etc. that can be smelted into new steel sheets, paper from packaging, office waste, etc that can be used in paper and carton production, and some kinds of plastic such as high-density polyethylene (HPDE) plastic used in water bottles that can be shredded in a process that results in plastic granules that can be used to model into new plastic products. 

More complex products can be deconstructed in a process that possibly identifies components that can be reused directly as spare parts (e.g. aircraft and cars are routinely stripped of parts that can be sold as spares) before the metal work is recycled. 
Also more complex products may contain valuable quantities of diverse metals (e.g. mobile phones often contain gold, silver, palladium, and platinum, as well as aluminium and copper) that require a more complex process to recover. However, consideration also needs to be given the risk for pollution arising from inappropriate methods being used for recovering metals from e-waste \cite{Zhang2012}.

Biowaste is an increasingly important topic in recycling. Often biowaste is regarded as something that needs to be disposed of as cheaply as possibly. Yet it can be a valuable resource. For example, after relatively simple treatment, human sewage can be used as a valuable form of fertiliser for forestry, or animal manure can be used in bioreactors to produce methane for electricity production. 

An interesting case study in biowaste repurposing arises from coffee. 
Various ways of exploitation of coffee wastes and by-products (coffee husks, pulp, immature and defective beans, coffee silverskin, and
spent coffee grounds) are presented and discussed in \cite{alves17}. These are motivated both by the huge amounts of coffee wastes generated each year but also by their valuable chemical compounds. Some notable examples of exploitation include using coffee husks and pulp as a supplement for animal feed, as organic fertilizers, for the production of enzymes, citric acid, biofuel or biogas, using spent coffee grounds as substrate for fungus growth or as adsorbent for the removal of heavy metals and using silverskin as an ingredient of cosmetics for skin hydration. Most of these types of exploitation require some chemical, physical or biological processes, such as the removal of caffeine and tannins from coffee husks before they are used for animal feed, the fermentation of husks or pulp for the production of enzymes or the extraction of coffee oil from defective beans. Some others, however, do not require any kind of modification; for example, coffee husks can be directly used as soil coverage. All of them, however, require knowledge of the chemical composition of the coffee wastes and the properties of the elements they consist of.

\subsection{Pharmaceuticals}

The development of new drugs is expensive, risky, and often takes years. An alternative that is increasingly appealing is repurposing of existing drugs to diseases other than those that they are licenced for \cite{Pushpakom2019,AMRC2017,NHS2020}. 
There are also many compounds that have been developed through pharmaceutical research that failed at some point in the development process but may be appropriate for another condition. An example is Viagra that was originally being developed for treating angina but had a side-effect that subsequently became its key selling point. 

Within the pharmaceutical industry, drug repurposing has become a hot-topic, in particular the development of computational techniques to identify candidates that are similar to existing drugs in terms of their biochemical or pharmacological characteristics, and so it may be viewed as a form of reuse exploitation.  An AI-based approach to drug repurposing is to analyse knowledge graphs constructed using NLP techniques (see for example \cite{Zhang2021}). However, knowledge graphs are a very simple formalism, and potentially, a deeper knowledge of repurposing would allow for more sophisticated methods for drug repurposing to be developed.

\subsection{Helping disadvantaged people}

There are many kinds of disadvantage experienced by people. An important kind of disadvantage is in healthcare in low income countries. With a lack of individual and government budgets to pay for equipment and resources that are taken for granted in higher income countries, alternative ways to equip and resource healthcare are required. One example is reuse of medical equipment by sending obsolete medical equipment from richer countries to poorer ones. This can be problematic because of lack of training, lack of necessary consumables, cost of use, cost of decommissioning, etc. \cite{Marks2019}. More problematic is the reuse of single use medical devices such as syringes even though medical guidelines prohibit such practices \cite{Wang2019}. So there is a need to consider the ramifications of an approach to repurposing in order to determine whether it is appropriate.

Nonetheless, repurposing can be valuable in finding cheaper alternatives for medical equipment and supplies. A good example is the UCL Ventura project for a medical device for providing breathing support for patients with severe respiratory problems (in particular arising from COVID-19 infection). Central to the remit of the design team was that the device could be manufactured in low income countries, and so existing designs for devices for providing continuous positive airways pressure were adapted to meet this pressing need \cite{Singer2020}. Healthcare practice can also be adapted for low and middle income countries. For example, diabetes is an increasing problem in many countries, but diabetes programmes to support behaviour change that are used in richer countries might have problems in low income countries (e.g. recommended diets might be difficult to afford or to provide in communities living in very poor quality and crowded accommodation that lacks water supply and reliable electricity, and provision of written advice and guidance might be challenging in communities with low levels of literacy)  \cite{Catley2020}. As another example, as a way to avoid the cost and difficulties of acquiring and operating expensive monitoring equipment for intensive care, inexpensive consumer devices for healthcare monitoring can be adapted for use in intensive care \cite{Turner2019}.

\subsection{Crisis management}

Crisis management is an unavoidable part of life. The COVID-19 pandemic has been a clear illustration of how many crises have arisen for individuals, organizations, and governments. We have seen how individuals have had to repurpose their homes into workspaces, teaching spaces, home gyms, etc. We have also seen many examples of how organisations have had to repurpose objects and processes to  address  public  requirements  or  economic  survival  (e.g.  exhibition  centres  becoming  isolation  hospitals,  vacuum  cleaner manufacturers  making  medical  ventilators,  schools  providing online learning). 

Crises come in many forms. For societies, they can come about from natural disasters (e.g. earthquakes/tsunamis, volcano eruptions, typhoons/hurricanes, flooding, wild fires, draught, etc). As a result, individuals and organizations need to adapt quickly. This can mean working out how to survive, and how to recover, using the means and resources available. How this is done depends on the circumstances. For instance, reaction to an earthquake, can depend on the degree to which existing facilities are damaged, what substitute resources are available (e.g. tents and materials for making temporary shelters), weather conditions (e.g. is there very hot or very cold weather, and is there rain or wind, that would make it difficult for people to survive in temporary shelters?), what infrastructure has survived (e.g. are there roads open to enable emergency help to arrive, is the telecommunications network working, is there potable water from the water supply network, does the sewage system work, etc?). Problems with any of these needed resources can mean that alternative resources need to be repurposed. For instance, if the water pumping system is broken, is there an alternative source of potable water such as a well in the area, and are there sufficient clean containers to be able to carry that water to where people are sheltering?



\section{Discussion}

There is a need for a general theory of repurposing, considering both substitution and exploitation. This would map out key parameters and steps in the process of repurposing such as: problem definition, identification of candidate solutions, and analysis/evaluation of solutions. For substitution, these phases would be based on the features or properties of the original, and the context (including constraints) of the substitution. For example, if we lack a chair, the properties of a wooden box (shape, stability, size, etc) might make it a candidate substitute, but for a dinner party context, aesthetic constraints might render it unacceptable. Complex situations would require complex formalisation. Emergency accommodation after an earthquake requires consideration of the weather conditions, how long shelter is required for, security needs, etc. The general theory would consider how candidate solutions can be identified with predictions from statistical models (e.g. node2vec embeddings), as well as abductive \cite{dickens14,kakas92,turliuc16} and argumentative \cite{atkinson17,besnard2008,flouris19} reasoning. Candidates could then be reasoned about in terms of mitigations and consequences, and given preference order based on specifications or heuristics~\cite{bothe13,grant17}.
So our claim is that, based on ideas from AI, a general theory of repurposing can be developed  to better understand this key human ability,
and this can then be used to engineer a set of AI methods and tools that can effectively and viably support repurposing in domains including cooking, DIY, recycling, crisis management, and helping disadvantaged people.


\subsection{Academic literature on the theory of repurposing}

Within some academic spheres there is a substantial literature on repurposing or related topics such as recycling, reuse, etc. with a focus on specific topics such as drug repurposing in pharmaceutical research, medical equipment reuse or repurposing for low income countries in healthcare research, biomass recycling in the environmental technology research, plastics recycling in material science research, etc. These lines of research provide detailed consideration of the specific problems and solutions, but they do not consider a general theory of repurposing, and hence they may miss the potential to develop general purpose tools to support the process.

Other academic fields touch on the nature of repurposing as part of investigating issues within that field, for example how repurposing can be used for understanding aspects of education \cite{Wardle2012}, and how government policy on environmental planning can be updated by repurposing \cite{Bishop2020}.
Furthermore, concepts related to repurposing appear in a number of academic disciplines, e.g. analogical reasoning in philosophy, psychology and AI, transfer in machine learning, metacognitive reasoning in pedagogy, drug repositioning in medicine, autograft in surgery, exaptation in evolutionary biology and appropriation in art.

Nonetheless, academic study of the general process of repurposing has only recently begun despite the importance of repurposing. The topic has been investigated from a historical and a cognitive perspective \cite{jones2013},
and from the perspective of how products can be designed with repurposing as a requirement \cite{Aguirre2006}, 
but neither of these address the need for a general theory for how we should go about repurposing.

There is one previous proposal for using a formalism to capture features of objects (namely, shape, material, and role of the object) and then reason with that knowledge to identify alternative uses \cite{olteteanu16}. 
For example, a cup can be defined in predicate logic as having a shape with high convexivity, and being made of ceramic, and for being used for drinking from.  
By using logic-based formalization, some objects can be substituted by others based on their definitions. 
But to understand more sophisticated kinds of substitution, and for considering other kinds of repurposing such as exploitation, we need to acquire and use more comprehensive knowledge about objects and services,  and consider both static and dynamic aspects of them. Furthermore, this knowledge needs to be automatically learned. 

More generally, there is research in analogical reasoning which has some relationship with understanding the reasoning behind repurposing. In AI, analogical reasoning is intended to capture how humans are able to understand new situations, and solve new problems, by identifying commonalities with previous situations and problems. So analogical reasoning involves identifying the key features involved as well as key differences \cite{Hall1989}. Even though it has long been investigated in AI, it has proved to be a challenging kind of reasoning to formalize in a robust and scalable way. A related topic is case-based reasoning which assumes a repository of cases, and a method for retrieving cases based on specific query criteria (e.g. retrieving cases of patients with specific combinations of symptoms that a new patient has, and the retrieved case can then be used to help diagnose and treat the new patient). Whilst case-based reasoning could be applied to the problem of supporting repurposing in specific domains, it does not offer us an underlying theory for repurposing, and it does not allow us have richer formalisms for representing and reasoning with the nature of objects and processes.

\subsection{Towards a general theory and tools for repurposing}

In our review we have argued that repurposing occurs in many spheres of life from the everyday through to the exceptional crisis situation. We have also argued that there are common features in repurposing across different domains, and that this could benefit from formalization as a general theory. 

In order to develop a general theory, we see that
there are a number of important strands of AI research that could be harnessed, and then tools for supporting repurposing could be based on AI technology. We briefly discuss some of these strands in the remainder of this section. 

\begin{itemize}

\item For pattern analysis, NLP could be useful for analysing corpora of examples for identifying patterns for repurposing. For example, if we collect a large corpus of cooking recipes, we can use text analysis tools to extract ingredient lists and, via statistical analysis, identify which ingredients are interchanged between different versions of the same, or similar, recipes. Also semi-supervised machine learning (ML) methods could be used to associate ingredients with classes (synsets) based on the usage of ingredients in recipes. The structure of a rich ontology can be extended and exploited in a number of ways. This includes representation learning approaches that \emph{embed} each entity as a vector of numbers that capture its properties (e.g. node2vec \cite{grover2016}), and have the potential to discover unnamed properties that help explain similarities of use, and thus support graph link prediction approaches to improve ontology coverage.

\item For property analysis, we can use ontologies to model information about the relevant entities of a domain, which may be derived from different sources and may be available in different formats, in a common way. For example, the literature on food science has good coverage of such classes (e.g. www.ifst.org). Recent developments in knowledge graphs can also be used for this purpose. For example, a recently developed knowledge graph, which contains structured (RDF) data about recipes and the nutritional value of ingredients is FoodKG (\url{foodkg.github.io/}). Some common substitutions that we can do in cooking can also be obtained from \url{www.foodsubs.com/} (for example, \url{www.foodsubs.com/Onionsdry.html} has a list of substitutes for onions, which include garlic). For the recycling case study, a useful resource could be this: \url{www.sciencedirect.com/science/article/pii/S2590238519303881.} More generally, ML and automated reasoning technologies need to be harnessed. This includes the development of methods for acquisition of knowledge using ML techniques \cite{muggleton99,turliuc16}, and for reasoning using knowledge-based technologies, such as description logic based systems for resource properties and features \cite{antoniou07}

\item For mechanistic analysis, we can draw on knowledge representation and reasoning.  Commonsense reasoning is particularly important in intelligence (see Table~\ref{table:commonsense}). There is a range of approaches in AI for such modelling including logic programming and event calculus~\cite{miller2002}. Commonsense reasoning is a core theme in AI \cite{brewka1991,mueller06,davis17}, and will be critical in developing systems for supporting repurposing. Commonsense reasoning frameworks can be harnessed for contextualising repurposing situations \cite{alrajeh13,dasaro20,diakidoy14,dickens14,kakas11,ma13,mueller06}. 
Commonsense reasoning is particularly important for moving beyond repurposing derived from pattern matching methods to repurposing based on richer knowledge and automated reasoning, e.g.\ supporting mitigations and explanations. 
So a key dimension to developing a general theory is to understand the types of commonsense reasoning and knowledge that arise in repurposing, and how these might be refined. E.g, in crisis management, there is often the need to substitute for basic human needs, but what is acceptable can vary between cultures. AI research in commonsense reasoning has not focused on culture differences, so the project may catalyse this study area.
Just look at how foods in one culture would not be acceptable as substitute even in extreme circumstances. 

\end{itemize}

We regard this work as forging a new domain of enquiry within AI. We hope that other AI practitioners will  participate in the wider research objectives and that numerous downstream stakeholders will recognise the concomitant benefits of study in this area. We will make our case studies and AI tools available via a website to enable further research  by other groups.

\begin{table*}
\begin{center}
\fbox{\begin{minipage}{16cm}
{\Large \bf What is commonsense reasoning?}

\vspace{5mm}

Commonsense reasoning is an innate ability of humans. 
It enables us to perceive, understand, judge, and operate in, the world around us in diverse spheres including: 
\begin{itemize}
\item {\bf How the physical world works} i.e. a naive physics (e.g. explaining why a vase breaks when dropped on the floor);
\item {\bf How people relate to events} (e.g. explaining the emotions of a colleague when they just get a proposal rejected);
\item {\bf How people operate in societies} (e.g. process to pay for a meal, how to get a credit card, etc);
\item and {\bf How organizations operate in societies} (e.g. how companies get funding, how companies can buy other companies, how laws get made in a country).
\end{itemize}
In each of these spheres, the aim of research into commonsense reasoning is to capture some of the intuitive understanding and behaviour that people normally display rather than develop academic models of physics, psychology, economics, or society. Obviously, this is an open-ended challenge, but in the short-term we can aim for capturing modest but useful fragments of this cognitive ability for specific applications, and use this as a stepping stone to the longer-term goal of developing more sophisticated commonsense reasoning. 

\vspace{5mm}

Cognitive activities rely on commonsense reasoning.
Consider disambiguating sentences to determine the most likely meaning (for example, from the sentence {\em John walked the baby round the park in the pram}, we can infer {\em John walked round the park}, but  not that {\em The park is in the pram};
vision understanding when viewing complex or noisy images to determine the most likely components of the image; 
decision-making when faced with some options and commonsense filters out inappropriate or non-viable options; 
risk analysis when using commonsense to predict what might go wrong in a particular scenario;
or problem solving by using commonsense solutions. 

\vspace{5mm}

So commonsense reasoning is important in understanding what we hear, see, touch and taste, and thereby it helps us understand the world around us. We use it to make sense of the world, it helps us identify opportunities and risks, and it helps us to solve problems and to make decisions. It is the key ingredient in our intelligence for making us flexible, robust, adaptable, and creative, and it is essential in making us proactive. It is also essential in helping us work to together with others. 
\vspace{5mm}
\end{minipage}}
\end{center}
\caption{\label{table:commonsense}Explanation of commonsense reasoning}
\end{table*}

\section*{Acknowledgements}

The authors are grateful to the Leverhulme Trust for supporting the project ``Repurposing of Resources:  from Everyday Problem Solving through to Crisis Management" which will run from January 2022 to December 2024.



\end{document}